\documentclass[10pt]{article}
\usepackage[letterpaper,margin=1.1in]{geometry}
\usepackage[T1]{fontenc}
\usepackage[utf8]{inputenc}
\usepackage{newtxtext,newtxmath}
\usepackage{graphicx}
\usepackage{booktabs,array,calc}
\usepackage{amsmath}
\usepackage{algorithm}
\usepackage{algpseudocode}
\usepackage{microtype}
\usepackage[hidelinks]{hyperref}
\providecommand{\tightlist}{\setlength{\itemsep}{0pt}\setlength{\parskip}{0pt}}
\title{\bfseries The World Model Remembers, the Actor Forgets:\\
Dream Rehearsal for Continual Model-Based RL}
\author{Gurp Nijjer\\ Quantegra Research, Canada\\ \small\texttt{gurp@quantegra.ca}}
\date{July 2026}

\begin{document}
\maketitle

\begin{abstract}
Model-based reinforcement-learning agents of the DreamerV3 family forget catastrophically when
trained on task sequences, even when an unbounded replay buffer preserves every earlier
experience. We ask a question the continual-RL literature has assumed an answer to but never
measured: \emph{which component forgets?} Under never-clear replay, pre-registered
component-level probes ($n{=}3$ seeds throughout) show that the world model retains
essentially everything measurable about old tasks --- reward discrimination (retention ratio
$\approx 1.0$), value estimates, and termination structure --- while the actor's behavior
collapses. Forgetting in this regime is a channel problem, not a memory problem. We
demonstrate this by intervention: with the world model frozen and identical imagined rollouts,
reinforcement learning in imagination fails to recover a lost skill (0/3 seeds), while
supervised self-imitation on the world model's own \emph{graded dreams} recovers it on 3/3
seeds with zero environment interaction. Interleaved during training, this \emph{graded dream
rehearsal} yields a task-label-free, parameter-constant continual learner: 3/3 four-task
chains retained where plain replay passes 0/3, 3/3 eight-task chains, and consistent gains
over matched real-episode cloning (paired difference $+0.13$, bootstrap 95\% CI
$[0.07, 0.24]$, complete seed separation). The dream-grading step is load-bearing: we
characterize two scoring failure modes, provide an offline selection gauge that caught both
before they contaminated results, and give a realized-first grading rule that closes them.
All experiments were pre-registered with committed protocols; every refuted hypothesis is
reported.
\end{abstract}

\vspace{0.4em}
\noindent Code, pre-registration trail, and run data:
\url{https://github.com/gurpnijjer/dream-rehearsal}\\
Archived: \href{https://doi.org/10.5281/zenodo.21462836}{doi:10.5281/zenodo.21462836}

\section{Introduction}

An agent that learns task B by destroying its competence at task A cannot learn continually,
and reinforcement-learning agents do exactly this. The standard remedies treat the network as
the thing to protect: regularize its parameters (EWC~\cite{kirkpatrick2017overcoming} and
successors), grow it (progressive networks~\cite{rusu2016progressive}, per-task heads), or
replay its past data (CLEAR~\cite{rolnick2019experience}; reservoir buffers). In model-based
RL the replay remedy has a natural home --- the world model trains on buffered experience ---
and the resulting recipe, a Dreamer-class agent with a never-clear buffer, is the field's
current working answer~\cite{kessler2023effectiveness,yang2024augmenting,alyahya2026arrow}.

That line of work carries an assumption stated explicitly but, to our knowledge, never
tested. WMAR~\cite{yang2024augmenting} writes: ``We hypothesise that maintaining the world
model's accuracy across different environments will help preserve performance.''
ARROW~\cite{alyahya2026arrow} infers world-model fidelity backward from behavioral metrics.
The premise --- protect the world model and the policy will follow --- has not been checked
at the component level, because the standard evaluation measures only episodic return, which
cannot distinguish a forgotten representation from a forgotten behavior.

We checked. The first half of the premise is right: under never-clear replay the world model
retains essentially everything we can measure about old tasks. The second half is wrong: the
actor forgets anyway. The training signal for old tasks --- accurate rewards, calibrated
values, intact dynamics over old-task states --- is present inside the agent's own
imagination throughout training, and the policy decays regardless. Forgetting, in this
regime, is not a memory problem. It is a channel problem: policy-gradient learning fails to
convert an intact signal into retained behavior, and we show a supervised channel converts
the same signal successfully.

A note on scope, stated here rather than left to the limitations: everything in this paper is
established on MiniGrid chains with a 17M-parameter agent. We chose this scale deliberately ---
it is the scale at which every component can be probed, frozen, checksummed, and re-read
per-seed, which is what a localization claim requires. The contribution is the phenomenon, its
component-level localization, and the mechanism that exploits it; whether the pattern persists
in larger domains is an open empirical question we are actively pursuing and do not claim
here.

This paper makes four contributions, in decreasing order of claimed novelty:
\begin{enumerate}\tightlist
\item \textbf{A component-level localization of forgetting} in world-model agents (\S4):
representations, reward heads, and critics survive under replay; only the actor's behavior
decays. This contradicts the untested premise underlying replay-based continual MBRL and
reframes what needs fixing.
\item \textbf{A channel-isolation result} (\S5): with a frozen world model and identical
imagined data, RL-in-imagination fails to recover a lost skill (0/3 seeds) where supervised
self-imitation succeeds (3/3, zero environment steps). The instability is the
policy-gradient channel itself.
\item \textbf{Graded dream rehearsal} (\S6): interleaved supervised self-imitation on
world-model-graded imagined rollouts of prior tasks during sequential training. One actor,
no task labels at any time, no parameter growth, ${\sim}15\%$ compute overhead at four
tasks. It converts the isolation upper bound into a mechanism --- 3/3 four-task and 3/3
eight-task chains retained, against 0/3 for the plain-replay recipe --- and consistently
outperforms matched real-episode cloning.
\item \textbf{Grading is load-bearing} (\S7): we characterize two failure modes of
imagined-trajectory scoring (post-terminal contamination; promises outbidding realized
successes), give a realized-first rule that closes both --- an imagination-safe port of the
self-imitation advantage gate~\cite{oh2018self} --- and introduce an offline gauge that
measures selection quality directly, which caught both failure modes before they cost
environment interaction.
\end{enumerate}

Every experiment was pre-registered before running --- protocols, pass bars, and
interpretation matrices committed to version control, with off-machine timestamped archives
--- and refuted hypotheses are reported in the main text, including two scoring bugs our own
gauges caught. We consider the gauges part of the method (\S7), and the transparency part of
the evidence.

\section{Related Work}

\textbf{Replay-protected continual MBRL.} Continual-Dreamer~\cite{kessler2023effectiveness},
WMAR~\cite{yang2024augmenting}, and ARROW~\cite{alyahya2026arrow} form the line whose premise
\S4 measures: a Dreamer-family agent whose world model trains on a retained (plain,
augmented, or distribution-matched) replay buffer, with continual competence attributed to
preserved world-model accuracy. Our plain-replay baseline is Continual-Dreamer's recipe at
our scale. ARROW is the closest concurrent method in setting, and it is orthogonal by our own
localization: ARROW protects the world model's training distribution with distribution-matched
\emph{real} replay, and \S4 shows that under never-clear replay the world model is already
the protected component. The actor is the component that forgets, and no method in this line
rehearses the actor from imagination.

\textbf{Generative replay and pseudo-rehearsal.} Deep Generative
Replay~\cite{shin2017continual} is the ancestral idea: train a generator of past data and
rehearse from it. Dream rehearsal is imagination-native --- no generator is trained, because
the world model already in the loop \emph{is} the generator --- and it grades trajectories
rather than replaying them. ReGen~\cite{govind2026world} (June 2026, concurrent) is
structurally the closest published mechanism: world-model pseudo-replay plus a single policy
trained by behavior cloning, in continual \emph{imitation} learning on a video-diffusion
model with instruction-conditioned replay. It is neither task-agnostic nor RL; we cite it as
convergent evidence for the mechanism family.

\textbf{Self-imitation.} The advantage gate of Self-Imitation
Learning~\cite{oh2018self} --- imitate a trajectory only when its realized return beats the
critic's estimate --- is prior art for \S7's grading rule on real data. Our contribution is
scoped to making that gate safe for \emph{imagined} data, where the critic's optimism is
unconstrained by real outcomes and where dreams do not stop at their own endings (\S7).

\textbf{Apparent counter-evidence.} Wo\l{}czyk et al.~\cite{wolczyk2024fine} show that
model-free fine-tuning irreversibly overwrites pre-trained capabilities. We reproduce the
distinction rather than contradict it: in their regime no mechanism exists that could
preserve anything (no world model, no dynamics objective, no retained data), and we observe
total representation overwrite in exactly that condition (\S4.2, no-replay bracket). Our
claim is regime-specific to replay-maintained world models --- the regime the continual-MBRL
literature actually operates in. The plasticity literature --- capacity loss under nonstationarity~\cite{lyle2022understanding},
primacy bias~\cite{nikishin2022primacy} --- documents degradation of the RL channel during
ordinary training and is mechanistically adjacent to \S5's channel finding; it does not,
however, measure backward transfer against an intact world model, which is the comparison our
setting isolates.

\textbf{Parameter isolation.} Frozen per-task modules with a task selector are the classical
isolation approach; \S4.4 constructs the task-agnostic version (frozen actor snapshots plus
a task router) as a measured reference point and shows why storing policies caps retention
at snapshot quality. CLEAR~\cite{rolnick2019experience} is the real-data ancestor of
\S6.2's cloning comparison, which we distinguish empirically rather than rhetorically.

\section{Experimental Setup}

\textbf{Agent.} DreamerV3~\cite{hafner2023mastering} (PyTorch reference implementation;
17M-parameter world model: CNN encoder, RSSM, reward/continuation/decoder heads;
1.8M-parameter actor; imagination horizon $H{=}15$). Greedy exploration; all hyperparameters
are the reference MiniGrid configuration unless stated (Appendix~\ref{app:hyper}).

\textbf{Tasks and chains.} MiniGrid~\cite{chevalier2023minigrid}, $64{\times}64$ RGB
observations. Core chain (\S\S4--7): DoorKey-5x5 $\rightarrow$ SimpleCrossingS9N1
$\rightarrow$ LavaGapS5 $\rightarrow$ MultiRoom-N2-S4 --- mixed episode horizons
(${\sim}10$ to ${\sim}80$ policy steps) and one lethal task, properties that turn out to
matter (\S7). The scale-up chain (\S8) appends four audition-gated tasks: LavaCrossingS9N1,
DistShift2, DoorKey-6x6, Unlock (eight tasks, six distinct mechanics, three lethal).

\textbf{Protocol.} Sequential phases. During phase $i$ the agent acts only in task $i$; a
shared never-clear buffer accumulates all experience (the Continual-Dreamer recipe at our
scale). Every 2{,}000 environment steps, every learned task is evaluated for 10
real-environment episodes. A phase advances when all learned tasks score ${\geq}0.6$ (mean
return) for 3 consecutive evaluation rounds, or at a 150k-step cap. Final retention is the
last-3-round mean per task; a seed passes if all tasks finish ${\geq}0.6$. $n{=}3$ seeds
everywhere; no per-seed tuning; every experiment's bars and interpretation matrix were
committed before launch. Because run-level nondeterminism is substantial (the same seed can
produce qualitatively different trajectories), we report per-seed values and dispersion
rather than pass counts alone.

\section{Which Component Forgets?}

\subsection{The phenomenon}
Two tasks, no replay: retention of task A after training task B falls from 0.96 to
0.0/0.27/0.12 across seeds. Four tasks \emph{with} unbounded replay: 0/3 chains pass, and
the casualty is the same task on every seed --- SimpleCrossing retains 0.35/0.51/0.25 while
DoorKey (${\sim}0.96$), LavaGap (${\sim}0.94$), and MultiRoom (0.79--0.82) survive. Replay
protects some tasks and not others; whatever replay protects, it is not uniformly ``the
policy.''

\subsection{Freeze bracket}
Freezing the actor's weights at task-A competence while the world model continues training:
retention 0.0/0.0/0.07 --- actor protection without representation stability is useless; the
latent space moves under the frozen head. Freezing the world model instead: the new task
never learns at all (the actor learns exclusively in imagination, and a frozen world model
only dreams task A) --- one seed, terminated early as a foregone conclusion and reported as
direction-finding only. The bracket eliminates both naive protection strategies and sharpens
the question: with replay keeping the world model trained on all tasks, what exactly is
still being lost?

\subsection{The world model retains everything we can measure}
We probed each component of the final chain checkpoint against every earlier phase
checkpoint (deterministic posterior-mode encoding; $n{=}3$; pre-registered with the
prediction that the reward head would show selective forgetting --- a prediction the data
refuted):
\begin{itemize}\tightlist
\item \emph{Reward knowledge.} Discrimination $D$ = (mean predicted reward at true success
steps) $-$ (at zero-reward steps) on each task's held episodes. The retention ratio
$R = D_{\text{final}}/D_{\text{own-phase}}$ for the always-forgotten task:
\textbf{0.99/1.06/1.01}. The reward head never forgets; on one seed it improves with
continued replay.
\item \emph{Off-distribution check.} The same discrimination on the degraded policy's own
final-era episodes --- the states a drifted actor actually visits, including outright
failures: $D = 0.87$ (vs.\ 0.90 on curated episodes), with zero false reward on failure
trajectories. The signal is intact where re-teaching needs it, not just where replay
polishes it.
\item \emph{Values.} Critic means on old-task states remain high and rise across phases
(0.84 $\rightarrow$ 0.91).
\item \emph{Termination model.} Continuation-head discrimination at true terminal steps:
0.95--1.0 at every checkpoint, including immediately after the shortest task phase.
\item \emph{Representations, with a caveat that matters.} A frozen-action-margin probe
shows old-task action preferences under re-encoded latents degrade (margin $-0.25$ at run
end vs.\ $+4.5$ and $+2.7$ for stabler tasks): the latent space \emph{does} drift.
Precisely: co-trained heads track the drift; frozen heads decay under it; and the actor ---
co-trained, but through the RL channel --- decays too. ``The world model remembers'' means
replay-maintained knowledge, not a frozen latent geometry.
\end{itemize}

A scope note: this is \emph{replay-maintained} memory. Without replay, representation
overwrite is total (\S4.2). The claim is regime-specific and consistent
with~\cite{wolczyk2024fine} (\S2). In the regime the continual-MBRL literature operates in,
the premise ``the world model is what needs protecting'' has it backwards: the world model
is the component replay already protects.

\subsection{The isolation reference}
For completeness we built the multi-head upper bound made task-agnostic: frozen per-task
actor snapshots plus a nearest-centroid task router over encoder embeddings (no task labels
at evaluation). With correct routing (measured ${\sim}1.00$ per task offline), the oracle
read passes 5/5 seeds --- but the honest routed read passes 3/5, entirely because the
hardest task's \emph{stored} policy delivers only $0.62 \pm 0.13$ against a 0.6 bar; two
seeds flipped verdicts in opposite directions between two reads of the same trained agent.
Storing policies caps retention at snapshot quality and converts the pass/fail question into
a coin flip at the bar. Isolation is a reference point, not a fix. (This experiment also
surfaced an episode-ordering bug in our router evaluation, whose diagnosis --- an offline
exact replication of the corrupted metric --- set the pattern for the selection gauges of
\S7; the corrected re-read is the number reported here, with both reads in the artifact.)

\section{Recovery from the World Model Alone}

If \S4 is right, the actor should be re-teachable from the world model alone. We froze the
entire final world model of each four-task run (encoder, RSSM, all heads; parameter
checksums asserted unchanged), selected the always-forgotten task's episodes from the buffer
by phase window (selection purity ${\geq}80\%$, verified by routing), and re-trained only
the drifted live actor on imagined rollouts from those states. Two teachers, same starts,
same budget (20k updates), zero new environment steps, real-environment evaluation every
500 updates (Figure~\ref{fig:recovery}):

\begin{figure}[t]
\centering
\includegraphics[width=0.85\linewidth]{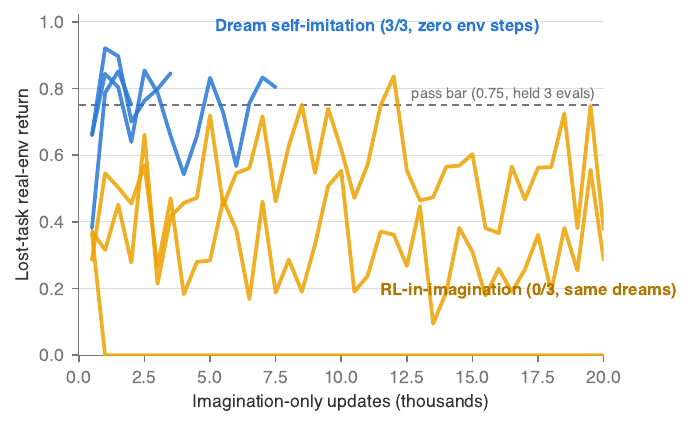}
\caption{\textbf{The recovery race.} Real-environment return on the forgotten task during
imagination-only re-teaching; three seeds per arm, identical frozen world model and imagined
data. Supervised self-imitation on graded dreams (blue) clears the 0.75 recovery bar on 3/3
seeds within 2{,}000--7{,}500 updates; RL-in-imagination (orange) thrashes for 20{,}000
updates and passes 0/3, with one seed collapsing to zero. Returns are raw environment return
in $[0,1]$; both arms use the same frozen world model and identical imagined data, so the
failure of the orange arm is attributable to the RL channel alone.}
\label{fig:recovery}
\end{figure}

\begin{itemize}\tightlist
\item \textbf{RL-in-imagination} (the standard DreamerV3 actor--critic update on the frozen
model): \textbf{0/3}. One seed reaches 0.84 and oscillates back to 0.38; one collapses to
0.0 \emph{and destroys collateral tasks' behavior} (a co-trained task falls to 0.08); one
never exceeds 0.57.
\item \textbf{Dream self-imitation} (roll the current actor with sampling in imagination;
grade each trajectory with the frozen reward and value heads; behavior-clone the top 25\%):
\textbf{3/3}, at 2{,}000/3{,}500/7{,}500 updates --- $0.38{\rightarrow}0.85$,
$0.66{\rightarrow}0.92$, $0.66{\rightarrow}0.85$ --- above the frozen-head storage ceiling
of \S4.4, with collateral tasks intact.
\end{itemize}

Same model, same knowledge, same dreams. The supervised channel converts the signal into
behavior; the policy-gradient channel thrashes. This was the pre-registered arm comparison's
strongest row (``only imitation passes $\rightarrow$ the instability is the RL channel''),
and it is why we frame forgetting in this regime as a channel problem. The world model is a
sufficient behavioral memory at this scale; what was missing was a stable way to read
behavior back out of it.

\section{Dream Rehearsal}

The recovery result suggests the fix: never let the actor drift in the first place
(Algorithm~\ref{alg:rehearsal}). After each task's phase ends, its buffered episodes are
kept as \emph{rehearsal starts}; from then on, after every 2{,}000 environment steps of
new-task training, the agent runs 50 dream-self-imitation updates per prior task ---
imagine from that task's states with the current (sampling) actor, grade each imagined
trajectory with the live reward/continuation/value heads (\S7), and behavior-clone the top
25\%. One live actor throughout. No task labels at training or evaluation, no frozen
policies, no router, no new parameters; the world model and critic train exactly as in the
plain recipe. Rehearsal adds ${\sim}15\%$ compute at four tasks (linear in task count).

\begin{algorithm}[t]
\caption{Dream rehearsal (one training chunk)}
\label{alg:rehearsal}
\begin{algorithmic}[1]
\Require world model $W$, actor $\pi$, critic $V$; never-clear buffer with per-phase episode
libraries $L_1,\dots,L_{k-1}$ for prior tasks; chunk size $C{=}2000$; updates per prior task
$K{=}50$; clone fraction $q{=}0.25$; horizon $H{=}15$
\State train $(W, V, \pi)$ for $C$ environment steps of the current task, exactly as in the
plain recipe
\For{$j = 1$ \textbf{to} $k-1$}
  \For{$K$ updates}
    \State sample a batch of states $s$ from $L_j$; encode with $W$
    \State imagine $H$ steps from $s$ with the sampling actor $\pi$
    \State score each imagined trajectory with the realized-first rule (Eq.~\ref{eq:score})
    \State behavior-clone $\pi$ on the top-$q$ fraction of trajectories
  \EndFor
\EndFor
\end{algorithmic}
\end{algorithm}

\subsection{Four-task result}
\begin{figure}[t]
\centering
\includegraphics[width=0.9\linewidth]{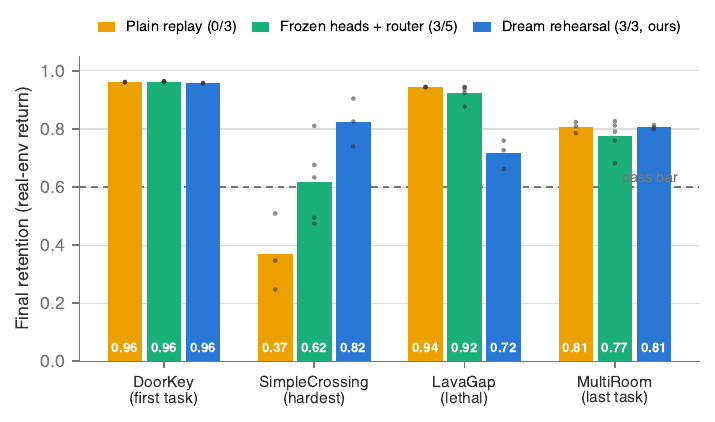}
\caption{\textbf{Final retention across the four-task chain}, three methods, per-seed points
over bar means, pass bar 0.6. Plain never-clear replay (orange) holds the easy tasks but
collapses on the hardest (SimpleCrossing, 0.37) --- 0/3 seeds pass all four. The
frozen-heads{+}router isolation reference (green) passes 3/5. Dream rehearsal (blue) passes
3/3 with the hardest task at 0.82.}
\label{fig:retention}
\end{figure}

\textbf{3/3 seeds pass all four tasks} (plain replay 0/3; isolation reference 3/5); per-seed
final retention in Table~\ref{tab:four}, method comparison in Figure~\ref{fig:retention}.

\begin{table}[h]
\centering
\small
\begin{tabular}{lccccc}
\toprule
Seed & DoorKey & SimpleCrossing & LavaGap & MultiRoom & Env.~steps \\
\midrule
1 & 0.959 & 0.905 & 0.760 & 0.814 & 136k \\
2 & 0.956 & 0.740 & 0.662 & 0.806 & 244k \\
3 & 0.958 & 0.826 & 0.727 & 0.799 & 130k \\
\bottomrule
\end{tabular}
\caption{Four-task chain, initial-scorer run. \S7's corrected grader lifts the one unstable
task's post-acquisition stability from 27/19/42\% to 95/100/67\% with all-task passes
preserved on all seeds; corrected-run per-seed means are 0.960/0.868/0.899/0.751,
0.961/0.751/0.944/0.800, and 0.964/0.825/0.784/0.768 (used in
Figure~\ref{fig:retention}).}
\label{tab:four}
\end{table}

The historically doomed task (SimpleCrossing) averages \textbf{0.824} --- above the
isolation ceiling's $0.62 \pm 0.13$ --- and stays above bar in \textbf{153/160 (96\%)} of
all post-acquisition evaluations. Two of three seeds finish the whole chain \emph{faster}
than the isolation architecture despite the rehearsal overhead, because no skill ever needs
relearning: acquisition of later tasks accelerates (LavaGap learns in 8--10k steps in-chain
on every seed). Retention converges to each task's own competence ceiling rather than a
common value --- which is what not forgetting should look like.

\subsection{Why dream at all? Comparison against real-episode cloning}
The buffer already stores the real episodes, actions included; cloning them directly
(CLEAR-style) is the obvious simpler alternative. We pre-registered this comparison
adversarially --- two of the four interpretation rows favor the simpler method, and
publication of the outcome in the main body was committed either way --- with the strongest
fair version: clone only \emph{competent} episodes (return $> 0.05$), matched
update-for-update to the dream schedule. A methods note: our first implementation of the
cloning arm silently misaligned features and actions by one step (the RSSM's action
convention lags the actor's decision), which \emph{destroyed} skills while cloning their own
successes --- an artifact flattering to our method, caught by an isolation gauge (cloning a
skill's own success must preserve it) and fixed before any comparison was read. With the
corrected update (Table~\ref{tab:ablation}):

\begin{table}[h]
\centering
\small
\begin{tabular}{lccc}
\toprule
Arm & $T_1$ per seed & Mean & All-four passes \\
\midrule
Dream rehearsal & 0.868 / 0.751 / 0.825 & \textbf{0.815} & 3/3 \\
Real-episode cloning (competent-filtered) & 0.630 / 0.678 / 0.743 & 0.684 & 3/3 \\
\bottomrule
\end{tabular}
\caption{Dream rehearsal vs.\ matched real-episode cloning; $T_1$ (SimpleCrossing) final
retention.}
\label{tab:ablation}
\end{table}

Real-episode cloning \emph{works}: it clears every bar and is cheaper per update, and we
report it as a viable baseline. But it retains consistently less: paired difference
$\mathbf{+0.131}$, bootstrap 95\% CI $\mathbf{[0.073, 0.238]}$ (10k draws), meeting the
pre-registered distinctness criterion (CI excludes zero, point estimate ${\geq}0.10$) ---
with complete seed separation (every dream seed exceeds every cloning seed). Imagination
contributes beyond the supervised channel; the natural candidates are fresh on-policy state
coverage from the current world model, and the gradability of imagined data (\S7). The
unfiltered cloning variant clones a strict superset of the filtered arm's data plus
pre-competence flailing and cannot plausibly exceed the filtered arm it would need to beat;
one seed of it completed under the corrected update ($T_1 = 0.649$ --- within the filtered
arm's per-seed range, below its mean, and below every dream seed), consistent with that
prediction; the remaining seeds were not run.

\section{Grading Imagined Trajectories Is Load-Bearing}

One systematic instability survived the four-task result: the lethal short-episode task
(LavaGap) wobbled after its phase --- above bar at the final read on all seeds, but clearing
the bar in only 27\%/19\%/42\% of post-acquisition evaluations. Pre-registered probes
(predictions committed; two of three refuted) convicted neither the rehearsal data (that
task's library was the largest and most competent of all tasks) nor the world model's
physics (termination discrimination 0.95--1.0 from the earliest checkpoint), but the
\textbf{scorer}: with ${\sim}10$-step episodes and a 15-step imagination horizon,
\textbf{99.7\%} of imagined trajectories terminate in-horizon, and a naive score (discounted
reward plus terminal value bootstrap) keeps scoring imagination \emph{past the dream's own
ending} --- post-terminal latents the model never trained on. Selection was near-random on
this task (AUC 0.49--0.56): 10--38\% of what rehearsal cloned were trajectories that walked
into lava, ranked there by post-terminal bootstrap noise. On long-horizon tasks, where
dreams rarely terminate in-horizon, the same scorer is clean --- which is exactly why only
the short-horizon task wobbled.

Fixing this exposed a second, opposite failure mode, caught \emph{offline, before any
environment cost}, by a selection gauge we now consider part of the method: generate sibling
dreams from banked starts, label each by its own imagined outcome, and measure each
candidate scorer's selection quality directly (AUC, top-quartile purity) rather than
inferring it from returns. Survival-weighting the rewards (with or without the exact
$\lambda$-return construction) \emph{inverts} selection on long-horizon tasks (AUC 0.004): a
dream that actually reaches the goal has its terminal reward and bootstrap suppressed, while
a dream that merely wanders keeps the critic's optimistic value --- promises outbid realized
successes. A recovery-referee run confirmed the inversion behaviorally (flat 0.0 where the
naive scorer had recovered the skill).

The rule that closes both failure modes is \textbf{realized-first grading}. Let $c_t$ be the
continuation head's probability at imagined step $t$, $p_t = \prod_{k<t} c_k$ the
probability the dream has \emph{reached} step $t$ (so nothing counts past termination),
$r_t$ the predicted reward, and $V$ the critic. With discount $\gamma$ and horizon $H$:
\begin{equation}
\label{eq:score}
\begin{aligned}
\text{realized} &= \sum_{t=0}^{H-1} \gamma^{t}\, p_t\, r_t,\\
\text{score} &= 10 \cdot \mathbf{1}[\text{realized} > 0.3] \;+\; \text{realized}
\;+\; \gamma^{H} p_H V(s_H).
\end{aligned}
\end{equation}
Trajectories that actually achieved reward in imagination outrank all value promises; the
termination-aware score orders within groups. The principle is not new on real data --- it
is the self-imitation advantage gate of Oh et al.~\cite{oh2018self}, $(R - V(s))_{+}$:
imitate only when realized return beats the critic's estimate. Our contribution is scoped to
making it safe for imagination, where the critic's optimism is unconstrained by any real
outcome and where dreams, unlike real episodes, do not stop at their own endings: the
reach-probability weighting, the termination-aware bootstrap, and the offline gauge. On the
gauge, realized-first scores AUC 1.0 and top-quartile purity 1.0 on both task profiles; in
the recovery referee it passes at 2{,}000 updates ($1.75\times$ faster than the naive
scorer); re-running the four-task chain with it lifts the wobbling task's stability to
\textbf{95\%/100\%/67\%} with every guardrail held (all-four passes on all seeds;
hardest-task mean 0.815). We note one adjacent null in the literature that our target
distinction predicts: reward-weighted \emph{episode selection for world-model training} does
not help and can cost plasticity~\cite[app.~D.8]{kessler2023effectiveness}. Selecting which
real episodes train the world model is a different object from selecting which imagined
trajectories the actor clones; our gauge measures the latter directly.

\section{Scaling to Eight Tasks}

Doubling chain length with four audition-gated additional tasks (each verified learnable
solo within the phase budget; candidates that could not bootstrap under greedy exploration
--- long-horizon sparse tasks --- were excluded and are reported as the acquisition frontier
of this study): DoorKey-5x5 $\rightarrow$ SimpleCrossing $\rightarrow$ LavaGap
$\rightarrow$ MultiRoom $\rightarrow$ LavaCrossing $\rightarrow$ DistShift2 $\rightarrow$
DoorKey-6x6 $\rightarrow$ Unlock. Same recipe, same constants; the only intentional change is task
count, though task identity necessarily varies with chain length --- we claim robustness of
the mechanism at doubled length, not a controlled scaling comparison. \textbf{All three seeds pass all eight tasks}, exceeding the pre-registered
${\geq}2/3$ bar (Table~\ref{tab:eight}).

\begin{table}[h]
\centering
\small
\setlength{\tabcolsep}{4.5pt}
\begin{tabular}{lcccccccc}
\toprule
Seed & DoorKey5 & SimpleX & LavaGap & MultiRoom & LavaCross & DistShift & DoorKey6 & Unlock \\
\midrule
1 & 0.964 & 0.943 & 0.947 & 0.759 & 0.861 & 0.960 & 0.915 & 0.861 \\
2 & 0.963 & 0.919 & 0.943 & 0.746 & 0.865 & 0.929 & 0.962 & 0.888 \\
3 & 0.901 & 0.876 & 0.942 & 0.754 & 0.794 & 0.959 & 0.962 & 0.740 \\
\bottomrule
\end{tabular}
\caption{Eight-task chain, per-seed final retention; pass bar 0.6 on every task.}
\label{tab:eight}
\end{table}

By the final phase, rehearsal runs 350 dream-imitation updates per 2{,}000-step chunk across
seven prior tasks --- still one actor, still no task labels, still no added parameters.

Two secondary reads matter more than the pass itself. \textbf{Consistency:} per-task
cross-seed ranges are ${\leq}0.07$ on seven of eight tasks (the exception is Unlock, range
0.15, still entirely above bar). The isolation reference of \S4.4 delivered
$0.62 \pm 0.13$ on its hardest task at \emph{half} this chain length, with two seeds
flipping verdicts between two reads of the same agent. At twice the tasks, rehearsal is
materially more stable than isolation was at half. \textbf{The predicted scale-failure did
not appear:} we pre-registered rehearsal dilution (early tasks sagging as a fixed per-task
budget spreads over more tasks) as the expected failure mode, with a stability-weighted
allocation rule as its remedy. No dilution is observed --- the incumbent quartet holds at or
above its four-task band --- so the remedy is not triggered and we do not build it.

\emph{Uncontrolled observation, explicitly not a result:} the historically doomed task reads higher at
eight tasks (0.943/0.919/0.876) than at four (0.868/0.751/0.825). We decline to interpret
this: the gap between arm means is the same magnitude as the four-task arm's own
seed-to-seed spread, the comparison crosses two different experiments (different downstream
tasks, rehearsal-pass counts, and budgets), and no matched-budget control was run. It is
recorded only so that a future controlled test --- same chain, varying only downstream task
count --- has a reason to exist.

Retention again settles at each task's own ceiling rather than a common value (easy tasks
${\sim}0.96$; MultiRoom ${\sim}0.75$, its band in every chain we have run; LavaCrossing
${\sim}0.86$).

\section{Limitations}

MiniGrid scale; a 17M-parameter world model; one task ordering per chain length; return bars
rather than normalized scores (we additionally report retention relative to each task's own
post-acquisition competence). $n{=}3$ with real run-level nondeterminism --- identical seeds
can diverge qualitatively --- so we report dispersion and bar-distance, not only pass
counts. We acknowledge $n{=}3$ is below the 5--10 common at scale; single-workstation compute
bounds it, and we compensate by reporting complete per-seed values for every headline
comparison, several of which (\S5, \S6.2) show complete seed separation, which no
aggregation can manufacture. Rehearsal cost grows linearly with task count; the observed acquisition speed-up
offsets it at $n{=}8$, but the crossover point is unmeasured. The buffer retains raw
experience: dream rehearsal removes stored \emph{policies}, not stored \emph{data} ---
replacing rehearsal starts with generated or bounded starts is future work. Tasks that
cannot bootstrap under greedy exploration within our budget mark the acquisition frontier of
this study; nothing here addresses exploration. Two scoring bugs shipped during development
and were caught by our own offline gauges before contaminating results; we report both
(\S4.4, \S7) and draw the methodological conclusion that selection gauges belong in the
method, not the appendix.

\section{Conclusion}

In replay-maintained world-model agents, forgetting is not a memory problem: the world model
retains what old tasks need, and the actor loses the behavior anyway. Once localized, the
failure has a direct fix --- read behavior back out of the world model through a supervised
channel, continuously, with graded dreams --- and the fix holds unanimously through
eight-task chains on a single workstation GPU. The grading rule, not the rehearsal schedule,
is where the difficulty lives; we ship the gauge that measures it. If the pattern holds
beyond gridworlds, the implication for continual MBRL is a reframing: protect less, and
re-teach from what the model already knows.

\subsection*{Reproducibility statement}
Every experiment was pre-registered: protocol, bars, and an interpretation matrix committed
to version control before launch (commit hashes in the artifact index), with off-machine
timestamped archives. All refuted hypotheses are reported. Code, run summaries, probe
artifacts, per-round evaluation traces, and the full pre-registration trail are available at
\url{https://github.com/gurpnijjer/dream-rehearsal} (archived at
\href{https://doi.org/10.5281/zenodo.21462836}{doi:10.5281/zenodo.21462836}). Total compute:
a single NVIDIA GB10 (Grace--Blackwell) workstation; the complete experimental record of this paper is
approximately three weeks of one GPU.

\appendix

\section{Protocol Details and Hyperparameters}
\label{app:hyper}

\begin{table}[h]
\centering
\small
\begin{tabular}{ll}
\toprule
Observation & $64 \times 64 \times 3$ RGB, partial egocentric view \\
World model & 17M parameters (CNN encoder, RSSM, reward/continuation/decoder heads) \\
Actor & 1.8M parameters; one-hot action distribution \\
Imagination horizon & $H = 15$ \\
Episode time limit & 256 environment steps \\
Train ratio & 512 (reference-implementation semantics) \\
Evaluation & every 2{,}000 env.\ steps; 10 real episodes per learned task \\
Phase advance & all learned tasks $\geq 0.6$ mean return for 3 consecutive rounds \\
Phase cap & 150k env.\ steps \\
Final retention & mean of last 3 evaluation rounds per task \\
Rehearsal schedule & 50 dream-imitation updates per prior task per 2{,}000-step chunk \\
Clone fraction & top 25\% of imagined trajectories per batch \\
Realized threshold & 0.3 (Eq.~\ref{eq:score}) \\
Exploration & greedy (no curiosity bonus) \\
Seeds & 1, 2, 3 for every graded experiment; no per-seed tuning \\
\bottomrule
\end{tabular}
\caption{Protocol constants. All remaining hyperparameters are the reference DreamerV3
MiniGrid defaults; the exact configuration files ship in the repository.}
\label{tab:hyper}
\end{table}

\noindent \textbf{Pre-registration.} Each experiment's protocol document (bars, arms,
interpretation matrix, and predicted failure modes) was committed to version control before
launch; amendments after first data are labeled as such in the released trail, with the
triggering observation disclosed. The released artifact includes every protocol, every
amendment, and both scoring-bug post-mortems.

\section{Per-Seed Data}
\label{app:seeds}

All per-run summaries (\texttt{chain\_summary.json}) and full evaluation traces
(\texttt{chain\_metrics.jsonl}; one row per evaluation round per task) for every run in this
paper --- plain-replay chains, freeze brackets, isolation-reference runs, recovery races,
both four-task rehearsal runs (initial and corrected scorer), the cloning-comparison arms,
and the eight-task chains --- are included in the released repository under
\texttt{results/}.

\section{Candidate Mechanisms for the RL Channel's Instability}
\label{app:mech}

Section~5 establishes \emph{that} the policy-gradient channel fails where supervised
imitation succeeds under identical conditions, not \emph{why}. We record the candidate
explanations consistent with our observations, explicitly as untested hypotheses.
\textbf{(1) Critic-optimism contamination.} \S7 documents that in imagination the critic's
optimism is unconstrained by real outcomes (value promises outbid realized successes when
scoring trajectories). The same optimism enters the actor--critic objective directly through
the bootstrapped return, so the RL channel chases promises no realized dream validates ---
consistent with the observed oscillation without convergence (one seed reaching 0.84 and
returning to 0.38). The supervised channel's targets are realized actions of graded
trajectories, with no bootstrap term. \textbf{(2) Update-magnitude interference.} One RL seed
destroyed a co-trained task's behavior (0.08) through the shared actor; the supervised channel
produced no collateral damage in any run. This is consistent with high-variance policy-gradient
updates propagating broadly through shared parameters. \textbf{(3) Relation to plasticity
findings.} Capacity loss~\cite{lyle2022understanding} and primacy
bias~\cite{nikishin2022primacy} document RL-channel degradation under nonstationarity during
ordinary training; our recovery setting shows failure even with a stationary, fully trained
world model, isolating the channel from the representation. Adjudicating among these requires
interventions we have not run --- variance clipping, critic-pessimism ablations, entropy
schedules --- and we list them as the natural next experiments.

\end{document}